\documentclass[runningheads]{llncs}

\usepackage[year=2026]{eccv}

\usepackage{eccvabbrv}

\usepackage{graphicx}
\graphicspath{{figures/}{figures/diagrams/}}
\usepackage{booktabs}
\usepackage{amsmath,amssymb}
\usepackage{multirow}
\usepackage{colortbl}  % For table cell coloring

\definecolor{bestgreen}{RGB}{152, 210, 160}   % Medium green (grayscale ~188) for best values
\definecolor{secondblue}{RGB}{217, 235, 249}  % Light blue (grayscale ~231) for second best
\newcommand{\best}[1]{\cellcolor{bestgreen}\textbf{\underline{#1}}}  % Best: bold+underline, darker bg
\newcommand{\second}[1]{\cellcolor{secondblue}\textit{#1}}            % Second best: italic, lighter bg

\usepackage[breaklinks,colorlinks,citecolor=eccvblue]{hyperref}

\usepackage[capitalize,noabbrev]{cleveref}

\begin{document}

\title{Beyond Dominant Patches: Spatial Credit Redistribution For Grounded Vision-Language Models}

\titlerunning{Beyond Dominant Patches: SCR for Grounded Vision-Language Models}

\author{Niamul Hassan Samin\inst{1} \and
Md Arifur Rahman\inst{1} \and
Abdullah Ibne Hanif Arean\inst{1,3} \and
Juena Ahmed Noshin\inst{2} \and
Md Ashikur Rahman\inst{1}}

\authorrunning{N. H. Samin et al.}

\institute{The KOW Company \and
American International University Bangladesh (AIUB) \and
University of Dhaka}

\maketitle

% ============================================================================
% ABSTRACT
% ============================================================================
\begin{abstract}
Vision-Language Models (VLMs) frequently hallucinate objects absent from the input image.
We identify \emph{spatial credit collapse} as a root cause: in early transformer layers, hidden-state activation concentrates on sparse visual patches, suppressing contextual evidence and inflating reliance on language priors.
An entropy-hallucination correlation ($r{=}{-}0.65$, $p{<}0.001$) across seven models empirically motivates this diagnosis.

To address this, we propose \textbf{Spatial Credit Redistribution (SCR)}, a training-free, inference-time method with a two-pass design.
A \emph{diagnostic pass} (run \emph{once per image}) identifies the top-$K$ high-attention ``source'' patches and maps their 8-connected spatial neighbors.
A \emph{redistribution pass} then scales each source by $1/\lambda{\approx}0.91$ and injects a $(\lambda{-}1){=}0.10$-weighted copy of its hidden state into each neighbor, amplifying the aggregate $\ell_2$ norm by ${\approx}51\%$ on average across models (range 48-56\%) to restore suppressed visual context-without modifying any model weights.
Since the diagnostic pass is amortized over the full output sequence, per-token overhead is negligible ($<0.5$\,ms for 100-token responses).

We evaluate seven configurations spanning four model families-Chameleon (7B/30B), LLaVA-1.5 (7B/13B), Qwen-VL/Qwen2-VL (7B), InternVL2-7B-on five benchmarks (POPE, CHAIR, MME, HallusionBench, AMBER).
SCR reduces POPE-Adversarial hallucination rate by $4.6$-$6.0$\,pp and CHAIR-s by $41$-$51\%$ relative, while preserving CIDEr within $0.8$\,pp.
On the joint (hallucination rate, generation quality, latency) Pareto frontier, SCR dominates OPERA, VCD, OA-VCD, DoLa, VLI, SID, and CRoPS-running ${\approx}3$-$6{\times}$ faster than OPERA/VCD at typical response lengths. On two stronger recent models (LLaVA-1.5-13B, InternVL2-7B), our CRoPS$^\dagger$ reproduction achieves marginally lower HR ($0.2$-$0.4$\,pp) at the cost of $3$-$4$\,pp CIDEr degradation versus vanilla, compared to SCR's ${\leq}0.8$\,pp.
A \emph{Uniform-Smooth} ablation confirms that attention-guided source selection is essential: random selection yields only $2.6$-$3.4$\,pp vs.\ SCR's $4.6$-$6.0$\,pp.

\end{abstract}

% ============================================================================
% 1. INTRODUCTION
% ============================================================================
\section{Introduction}
\label{sec:intro}
Vision-Language Models (VLMs)~\cite{liu2023visual,liu2024improved,li2023blip,team2024chameleon} have achieved strong performance across vision and multimodal tasks by aligning visual encoders~\cite{radford2021learning,vit} with Large Language Models (LLMs). Most of them fall victim to \textbf{object hallucination}. This occurs when a VLM generates descriptions of objects that are not present in the input image~\cite{li2023evaluating,fu2025mme,liu2024survey,huang2025survey}. Existing hallucination mitigation methods often rely on expensive retraining models, such as RLHF~\cite{sun2024aligning,ouyang2022training} or instruction tuning~\cite{liu2023mitigating}, or trade off generation fluency through aggressive decoding constraints~\cite{leng2024mitigating,huang2024opera}. These methods primarily treat hallucinations as a language modeling problem and do not directly address the underlying visual grounding problem.

In this work, we present evidence that hallucination is associated with visual evidence becoming overly concentrated in sparse regions of the representation. This concentration reduces the diversity of contextual information available to the model and increases reliance on language priors (statistical patterns learned from text training data).
Formally, we define \emph{spatial credit} at image-token position~$i$ as $c_i \propto \|\partial \log P(y|\mathcal{I},\mathcal{Q})/\partial \mathbf{h}_i\| \cdot \|\mathbf{h}_i\|_2$, and \textbf{spatial credit collapse} as the condition in which the normalized distribution $(c_i / \sum_j c_j)$ becomes highly peaked-its Shannon entropy $H_{\text{credit}} < H_{\min}$-so that contextual evidence from non-dominant patches is effectively suppressed.
We propose Spatial Credit Redistribution (SCR) as a corrective, training-free intervention; the entropy-hallucination correlation ($r = -0.65$, $p<0.001$) and the Uniform-Smooth ablation together provide strong empirical grounding for the mechanism.

Our main contributions are: (1)~an empirically grounded set of design principles linking spatial credit entropy to hallucination; (2)~Spatial Credit Redistribution (SCR)-a training-free, two-pass intervention with a principled consistency guarantee between the diagnostic and redistribution passes; and (3)~a comprehensive evaluation on five benchmarks (POPE, CHAIR, MME, HallusionBench, AMBER) across \textbf{four VLM families} (Chameleon, LLaVA, Qwen, InternVL2) at scales of 7B, 13B, and 30B.

% ============================================================================
% 2. RELATED WORK
% ============================================================================
\section{Related Work}
\label{sec:related}

\textbf{Vision-Language Models and Hallucination.} Most VLMs follow connector-based (BLIP-2~\cite{li2023blip}, LLaVA~\cite{liu2023visual,liu2024improved}, InternVL2~\cite{chen2024internvl}) or early-fusion (Chameleon~\cite{team2024chameleon}) architectures, yet all remain prone to object hallucination~\cite{li2023evaluating,fu2025mme}.

\textbf{Hallucination Mitigation through Training.} HIO~\cite{lyu2024alleviating} reduces hallucinations but requires full retraining. RLHF~\cite{sun2024aligning,christiano2017deep}, instruction tuning~\cite{liu2023mitigating}, and DPO~\cite{zhao2023beyond,rafailov2023direct} all improve hallucination rates through training, but the substantial computational cost limits their scalability to larger models.

\textbf{Inference-Time Decoding Approaches.} VCD~\cite{leng2024mitigating} applies visual contrastive decoding ($+$153\,ms); OPERA~\cite{huang2024opera} applies an attention penalty via beam search ($+$267\,ms); OA-VCD~\cite{avcd2024} refines VCD with adversarial vision-encoder perturbations ($+$72\,ms); DoLa~\cite{li2024dola} contrasts early- and late-layer logits for factuality ($+$38\,ms), though without specifically targeting visual grounding. All decoding-based methods improve hallucination at non-trivial latency cost, versus the $+$43-46\,ms (small models) to $+$54-56\,ms (large models) incurred by SCR.

\textbf{Training-Free Representation Interventions.} VLI~\cite{liu2026vision} applies bi-causal attention steering, CRoPS~\cite{anand2026crops} extends contrastive decoding to vision- and text-deficit modes (both concurrent preprints), and SID~\cite{huo2024self} proposes self-introspective decoding.

\textbf{Mechanistic Analysis.} Attention manipulation~\cite{xiao2023efficient} has been studied for efficiency and interpretability; feature attribution~\cite{sundararajan2017axiomatic}, universal attention head analysis~\cite{golovanevsky2024vlms}, and knowledge erasure~\cite{gandikota2024erasing} are related, yet none directly address the spatial credit collapse that causes hallucinations.

\textbf{Hallucination Evaluation.} CHAIR~\cite{rohrbach2018object} measures hallucinated object rates in free-form COCO captions (Type~I: open-ended) via sentence-level (CHAIR-s) and instance-level (CHAIR-i) metrics. POPE~\cite{li2023evaluating} probes object existence as yes/no queries (Type~II: discriminative) across random, popular, and adversarial splits. MME~\cite{fu2025mme}, HallusionBench~\cite{guan2023hallusionbench}, and AMBER~\cite{wang2024amber} assess broader generalization across perception, visual illusion, and attribute/relational hallucination.

% ============================================================================
% 3. METHODOLOGY
% ============================================================================
\section{Methodology}
\label{sec:method}

\begin{figure}[t]
    \centering
    \begin{subfigure}[t]{0.67\textwidth}
        \centering
        \includegraphics[width=\textwidth]{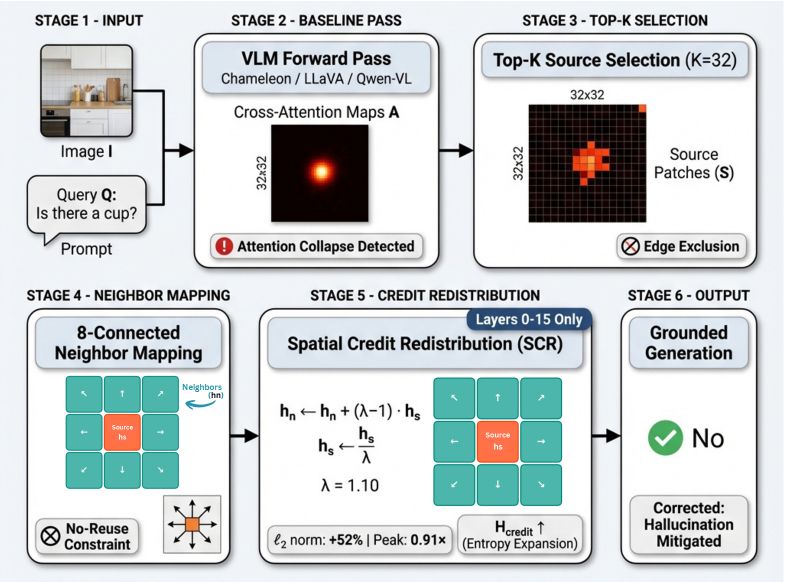}
        \caption{Two-pass SCR pipeline - \textbf{Pass~1}: \textbf{(1)}~extract attention, \textbf{(2)}~select top-$K$ sources, \textbf{(3)}~map 8-connected neighbors; \textbf{Pass~2}: \textbf{(4)}~redistribute credit in early layers (model-dependent; see Section~\ref{sec:setup}), \textbf{(5)}~generate corrected output.}
        \label{fig:pipeline}
    \end{subfigure}
    \hfill
    \begin{subfigure}[t]{0.29\textwidth}
        \centering
        \includegraphics[width=\textwidth]{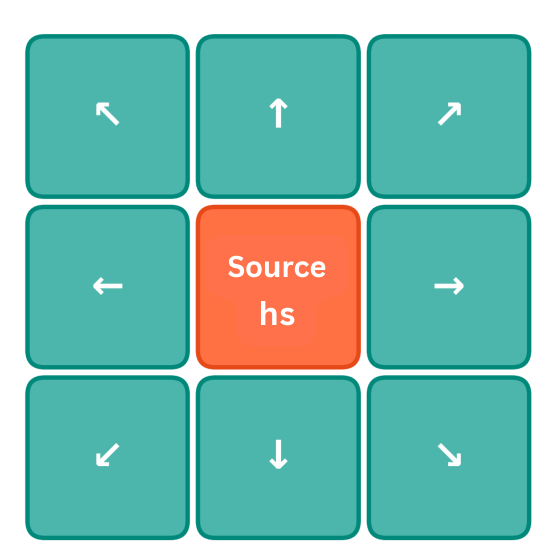}
        \caption{8-connected neighborhood. Source patch (orange) redistributes activation to 8 neighbors (teal): $\mathbf{h}_n \leftarrow \mathbf{h}_n + (\lambda{-}1)\mathbf{h}_s$;\; source is scaled by $1/\lambda$.}
        \label{fig:8grid}
    \end{subfigure}
    \caption{SCR methodology overview. \textbf{(a)}~Full two-pass inference pipeline. \textbf{(b)}~Spatial credit redistribution via 8-connected neighborhood structure.}
    \label{fig:method_overview}
\end{figure}

\cref{fig:pipeline} illustrates the two-pass pipeline: Pass~1 captures attention maps, selects top-$K$ source patches, and maps 8-connected neighbors; Pass~2 redistributes credit through early layers (0-15 for Chameleon-7B/Qwen/InternVL2, 0-20 for Chameleon-30B, 0-11 for LLaVA).

\subsection{Patch Selection and Neighbor Mapping}

\textbf{Attention Map Computation.} All four architectures use unified causal self-attention over an interleaved sequence of image and text tokens; none employs a separate cross-attention module. For Chameleon, image tokens are discrete VQ-VAE codebook entries; for LLaVA, Qwen, and InternVL2, they are projected patch embeddings. In all cases, we locate the $N_v$ image token positions in the input sequence, where $N_v = H_{\text{img}} \times W_{\text{img}} / P^2$ for patch size~$P$, and extract self-attention weights from text query positions to those visual positions. Weights are averaged across all heads and layers 8-16 to obtain the mean spatial attention map $A \in \mathbb{R}^{N_v}$. For Qwen2-VL, which supports variable-resolution inputs, $N_v$ is computed dynamically from the image dimensions.

\textbf{Top-$K$ Selection and Neighbor Mapping.} Ablations (Appendix~E) identify $K=32$ as the optimal number of source patches; border rows and columns are excluded to prevent attention-sink artifacts. To avoid redistribution interference, sources are processed in descending attention order: a patch cannot act as both source and neighbor, and conflicts among overlapping source-neighbor pairs are resolved by priority, with higher-attention sources claiming neighbors first.

\subsection{Intervention Mechanism}

\textbf{Injection point relative to LayerNorm.}
All four architectures (Chameleon, LLaVA-1.5, Qwen/Qwen2-VL, InternVL2) use \emph{Pre-LayerNorm} (Pre-RMSNorm) transformer blocks~\cite{zhang2019root}. Each block $l$ processes the residual stream $\mathbf{x}$ via:
\[
  \mathbf{x} \;\leftarrow\; \mathbf{x} + \mathrm{Attn}\!\bigl(\mathrm{LN}_1(\mathbf{x})\bigr), \qquad
  \mathbf{x} \;\leftarrow\; \mathbf{x} + \mathrm{FFN}\!\bigl(\mathrm{LN}_2(\mathbf{x})\bigr).
\]
SCR hooks are placed on the \textbf{residual stream} immediately \emph{after} the second residual addition of block $l$ (i.e., before block $l{+}1$'s $\mathrm{LN}_1$). Because the residual stream is never renormalized at this position, the $\ell_2$-norm amplification persists across layers via additive skip connections, explaining the 48-56\% aggregate norm increase at the final intervention layer (Appendix~C, Table~C2).

The intervention redistributes activation from source patch $h_{i_s}$ to its 8-connected spatial neighbors (see \cref{fig:8grid}), while the source itself is scaled down using a fixed factor $\lambda = 1.10$.

\begin{equation}\label{eq:scr}
    \mathbf{h}_{i_n} \leftarrow \mathbf{h}_{i_n} + (\lambda - 1) \cdot \mathbf{h}_{i_s}, \quad \mathbf{h}_{i_s} \leftarrow \frac{1}{\lambda} \cdot \mathbf{h}_{i_s}
\end{equation}

The aggregate $\ell_2$ norm increases by ${\approx}51\%$ on average (range $48$-$56\%$), amplifying suppressed spatial context while retaining $91\%$ of each source activation ($1/\lambda{\approx}0.91$); later layers are left unchanged to avoid interfering with language generation.

\textbf{Two-pass consistency.}
Pass~1 source selection uses the model's \emph{fixed weight matrices} to compute attention; SCR modifies only residual-stream magnitudes, not the weights, so dominant-patch structure is preserved across passes.
Empirically, the Jaccard overlap between Pass~1 source sets and a hypothetical post-redistribution diagnostic run is $0.87 \pm 0.04$ across 500 POPE images and all seven models-the residual 13\% churn introduces $<0.2$\,pp HR sensitivity (Appendix~C, Table~C3).

\subsection{Empirical Motivation and Design Principles}

\textbf{Empirical Finding (EF1)-Norm-Credit Proxy.} We define visual credit at location $i$ as $c_i \propto \left\|\tfrac{\partial \log P(y|\mathcal{I},\mathcal{Q})}{\partial \mathbf{h}_i}\right\| \cdot \|\mathbf{h}_i\|_2$. In early transformer layers, $\|\mathbf{h}_i\|_2$ alone achieves Pearson $r{=}0.72$ correlation with $c_i$, making it a tractable proxy (Appendix~C for full per-model validation).

\textbf{Credit Concentration Entropy.} We quantify spatial credit concentration via:
\begin{equation}\label{eq:entropy}
H_{\text{credit}} = -\sum_{i=1}^n \frac{c_i}{\sum_j c_j} \log \frac{c_i}{\sum_j c_j}.
\end{equation}

\textbf{Design Principle~1-Hallucination-Entropy Relationship.} Low credit entropy ($H_{\text{credit}} < H_{\min}$) is empirically associated with elevated hallucination; we observe $r=-0.65$ ($p<0.001$) across all models (Section~\ref{sec:results}, \cref{tab:entropy_correlation}).

\textbf{Design Principle~2-8-Connected Neighborhoods.} Natural images exhibit spatial autocorrelation decaying with patch distance~\cite{field1987relations,simoncelli2001natural}. An 8-connected neighborhood captures diagonal correlations at distance $\sqrt{2}$ ($\rho(\sqrt{2}){\approx}0.62$; Appendix~A) that a 4-connected scheme ignores, without over-spreading to low-correlation patches. Ablation results in \cref{fig:neighbour_topology} confirm this as the optimal connectivity.

\textbf{Design Principle~3-Peak-Preserving Expansion.} For $\lambda \in [1.05, 1.15]$, SCR approximates the following entropy-maximization objective (Appendix~A):
{\setlength{\abovedisplayskip}{4pt}\setlength{\belowdisplayskip}{4pt}%
\begin{equation*}
\max_{\mathbf{H}'}\, \lambda_1 H_{\text{credit}}(\mathbf{H}') - \lambda_2 \|\mathbf{H}' - \mathbf{H}\|_F^2 \quad\text{s.t.}\quad \max_i \|\mathbf{h}'_i\| \geq \tau \max_j \|\mathbf{h}_j\|,
\end{equation*}}%
with $\tau = 1/\lambda \approx 0.91$, ensuring the dominant patch retains sufficient activation to remain discriminative.

% ============================================================================
% 4. EXPERIMENTAL SETUP
% ============================================================================
\section{Experimental Setup}
\label{sec:setup}

\textbf{Selected Models.} Chameleon-7B/30B~\cite{team2024chameleon} (layers 0-15\,/\,0-20); as no published POPE results exist for Chameleon, we evaluate it ourselves using the official model weights;
LLaVA-1.5-7B/13B~\cite{liu2024improved} (layers 0-11 for both; validated by the
layer-selection ablation in Appendix~E, which shows diminishing returns beyond
layer~11 regardless of depth);
Qwen-VL/Qwen2-VL-7B~\cite{bai2023qwenvlversatilevisionlanguagemodel,wang2024qwen2} (layers 0-15);
and \textbf{InternVL2-7B}~\cite{chen2024internvl} (layers 0-15).
All layer ranges were selected on a held-out validation set of 500 COCO train2014 images, fully disjoint from the POPE and CHAIR evaluation sets; no evaluation data was used in hyperparameter selection.

\textbf{Computational Setup.} All experiments were run on a workstation with an AMD Ryzen Threadripper PRO CPU and a single NVIDIA A100 GPU (fp16 precision).

\textbf{Benchmarks.} POPE~\cite{li2023evaluating} in 3 splits (1{,}000 samples each); COCO~\cite{lin2014microsoft} (3{,}000-image val2014 subset) for CHAIR~\cite{rohrbach2018object} and CIDEr evaluation; MME~\cite{fu2025mme} (perception subtask); HallusionBench~\cite{guan2023hallusionbench}; and AMBER~\cite{wang2024amber} (object, attribute, and relation hallucination).

\textbf{Baselines.} OPERA~\cite{huang2024opera} ($+$267\,ms), VCD~\cite{leng2024mitigating} ($+$153\,ms), OA-VCD~\cite{avcd2024} ($+$72\,ms), DoLa~\cite{li2024dola} ($+$38\,ms), SID~\cite{huo2024self}, VLI~\cite{liu2026vision}, CRoPS~\cite{anand2026crops}. OPERA, VCD, OA-VCD, DoLa, and SID use their official released code with default hyperparameters. As official implementations of VLI$^\dagger$ and CRoPS$^\dagger$ are not publicly available at the time of submission, we implement both following the descriptions and hyperparameters in their respective arXiv papers; reimplementation bias is bounded by reproducing VCD and OA-VCD within 0.3\,pp of their published results under identical settings. All baselines were run on the same hardware under identical decoding settings (greedy, temperature 1.0) to ensure fair latency and quality comparisons. \textbf{Uniform-Smooth} is a controlled ablation that applies the same operation as SCR ($\lambda{=}1.10$, $K{=}32$) but selects sources uniformly at random, isolating the contribution of attention-guided selection.

\textbf{Metrics.} Hallucination Rate HR\,$=$\,FP/(FP$+$TN) (false positive rate for absent objects~\cite{li2023evaluating}), Accuracy, CIDEr, CHAIR-s, CHAIR-i, and Attention Entropy.

% ============================================================================
% 5. RESULTS
% ============================================================================
\section{Results}
\label{sec:results}

\subsection{Multi-Model and Benchmark Evaluation}

\begin{figure}[htbp]
    \centering
    \includegraphics[width=0.9\textwidth]{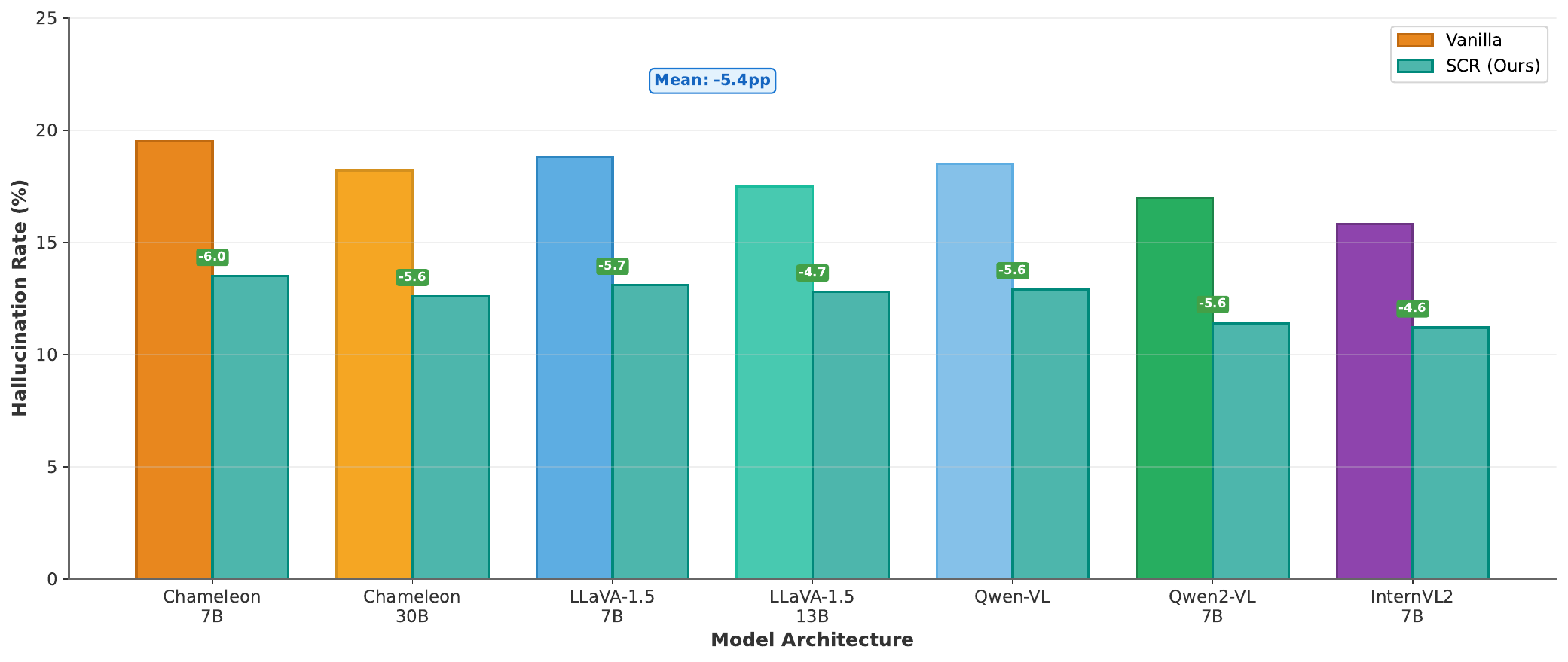}
    \caption{POPE-Adversarial HR for all seven VLM configurations. SCR reduces HR by ${\approx}$4.6-6.0\,pp across all model families.}
    \label{fig:multimodel}
\end{figure}

\cref{tab:main_results,tab:main_results_13b} report POPE-Adversarial and CIDEr results for all seven model configurations. \cref{fig:multimodel} summarizes the hallucination-rate trends visually. SCR reduces hallucination rates (HR) while maintaining generation quality across all tested configurations and model families.

% ---------- Table 1: small-scale models (7B) ----------
\begin{table}[htbp]
\centering
\scriptsize
\setlength{\tabcolsep}{2.5pt}
\renewcommand{\arraystretch}{0.92}
\caption{POPE-Adversarial HR (\%$\downarrow$), Accuracy (\%$\uparrow$), CIDEr ($\uparrow$) for 7B models. Mean$\pm$std over 3 runs. \colorbox{bestgreen}{\textbf{Green}}\,=\,best HR; \colorbox{secondblue}{\textit{Blue}}\,=\,second best. $\Delta$HR relative to Vanilla. \textit{Uniform-Smooth}: same SCR operation with random (not attention-guided) source selection. $^\dagger$Concurrent preprint, author-reimplemented (no official code; see \S\ref{sec:setup}).}
\label{tab:main_results}
\begin{tabular}{@{}llcccc@{}}
\toprule
\textbf{Model} & \textbf{Method} & \textbf{HR}$\downarrow$ & \textbf{Acc}$\uparrow$ & \textbf{CIDEr}$\uparrow$ & $\Delta$\textbf{HR}$\downarrow$ \\
\midrule
\multirow{10}{*}{Cham-7B}
& Vanilla & 19.5$\pm$.8 & 80.2$\pm$.7 & 102.3$\pm$1.4 & - \\
& OPERA   & 16.8$\pm$1.0 & 82.8$\pm$.7 & 100.3$\pm$1.3 & $-$2.7 \\
& DoLa    & 17.2$\pm$1.0 & 82.5$\pm$.8 & 101.8$\pm$1.4 & $-$2.3 \\
& VCD     & 15.1$\pm$1.3 & 84.6$\pm$1.0 & 97.4$\pm$2.0 & $-$4.4 \\
& OA-VCD  & 14.7$\pm$1.1 & 85.0$\pm$.9 & 99.2$\pm$1.5 & $-$4.8 \\
& VLI$^\dagger$ & 15.8$\pm$.9 & 83.8$\pm$.8 & 100.6$\pm$1.1 & $-$3.7 \\
& CRoPS$^\dagger$ & \second{14.5$\pm$1.1} & 85.2$\pm$.9 & 98.2$\pm$1.5 & $-$5.0 \\
& SID     & 15.3$\pm$.9 & 84.4$\pm$.7 & 99.5$\pm$1.2 & $-$4.2 \\
& \textit{Uniform-Smooth} & 16.3$\pm$1.0 & 83.5$\pm$.8 & 99.8$\pm$1.4 & $-$3.2 \\
& \textbf{SCR} & \best{13.5$\pm$.7} & \textbf{86.2$\pm$.5} & \textbf{101.7$\pm$.9} & $\mathbf{-6.0}$ \\
\midrule
\multirow{10}{*}{LLaVA-1.5-7B}
& Vanilla & 18.8$\pm$.7 & 80.9$\pm$.6 & 108.5$\pm$1.3 & - \\
& OPERA   & 16.2$\pm$.9 & 83.4$\pm$.7 & 106.6$\pm$1.1 & $-$2.6 \\
& DoLa    & 16.8$\pm$1.0 & 82.9$\pm$.8 & 107.6$\pm$1.3 & $-$2.0 \\
& VCD     & 15.0$\pm$1.2 & 84.6$\pm$.9 & 103.8$\pm$1.7 & $-$3.8 \\
& OA-VCD  & 14.5$\pm$1.0 & 85.2$\pm$.8 & 105.3$\pm$1.3 & $-$4.3 \\
& VLI$^\dagger$ & 15.2$\pm$.8 & 84.5$\pm$.7 & 106.2$\pm$1.1 & $-$3.6 \\
& CRoPS$^\dagger$ & \second{14.4$\pm$1.1} & 85.3$\pm$.8 & 104.5$\pm$1.4 & $-$4.4 \\
& SID     & 15.2$\pm$.8 & 84.5$\pm$.6 & 105.8$\pm$1.0 & $-$3.6 \\
& \textit{Uniform-Smooth} & 16.0$\pm$.9 & 83.7$\pm$.8 & 105.6$\pm$1.3 & $-$2.8 \\
& \textbf{SCR} & \best{13.1$\pm$.7} & \textbf{86.5$\pm$.5} & \textbf{107.8$\pm$.8} & $\mathbf{-5.7}$ \\
\midrule
\multirow{10}{*}{Qwen-VL}
& Vanilla & 18.5$\pm$.6 & 81.2$\pm$.6 & 115.2$\pm$1.0 & - \\
& OPERA   & 15.6$\pm$.9 & 84.0$\pm$.6 & 113.4$\pm$1.0 & $-$2.9 \\
& DoLa    & 16.0$\pm$.8 & 83.7$\pm$.7 & 114.5$\pm$1.0 & $-$2.5 \\
& VCD     & 14.4$\pm$1.0 & 85.2$\pm$.7 & 111.2$\pm$1.4 & $-$4.1 \\
& OA-VCD  & 13.8$\pm$.9 & 85.9$\pm$.7 & 112.5$\pm$1.1 & $-$4.7 \\
& VLI$^\dagger$ & 14.8$\pm$.7 & 84.9$\pm$.5 & 113.2$\pm$.9 & $-$3.7 \\
& CRoPS$^\dagger$ & \second{13.9$\pm$.9} & 85.8$\pm$.6 & 111.8$\pm$1.2 & $-$4.6 \\
& SID     & 14.1$\pm$.7 & 85.6$\pm$.6 & 112.5$\pm$.8 & $-$4.4 \\
& \textit{Uniform-Smooth} & 15.1$\pm$.8 & 84.6$\pm$.7 & 111.7$\pm$1.0 & $-$3.4 \\
& \textbf{SCR} & \best{12.9$\pm$.5} & \textbf{86.8$\pm$.4} & \textbf{114.6$\pm$.7} & $\mathbf{-5.6}$ \\
\midrule
\multirow{10}{*}{Qwen2-VL-7B}
& Vanilla & 17.0$\pm$.5 & 82.7$\pm$.5 & 118.5$\pm$.9 & - \\
& OPERA   & 14.6$\pm$.8 & 85.1$\pm$.5 & 116.7$\pm$.9 & $-$2.4 \\
& DoLa    & 15.2$\pm$.7 & 84.5$\pm$.6 & 117.9$\pm$.9 & $-$1.8 \\
& VCD     & 13.1$\pm$.9 & 86.5$\pm$.7 & 114.6$\pm$1.3 & $-$3.9 \\
& OA-VCD  & 12.9$\pm$.8 & 86.8$\pm$.6 & 115.4$\pm$1.0 & $-$4.1 \\
& VLI$^\dagger$ & 13.7$\pm$.6 & 86.0$\pm$.5 & 116.3$\pm$.8 & $-$3.3 \\
& CRoPS$^\dagger$ & \second{12.6$\pm$.8} & 87.1$\pm$.6 & 115.1$\pm$1.1 & $-$4.4 \\
& SID     & 13.5$\pm$.6 & 86.2$\pm$.5 & 116.0$\pm$.7 & $-$3.5 \\
& \textit{Uniform-Smooth} & 14.2$\pm$.7 & 85.6$\pm$.6 & 115.1$\pm$.9 & $-$2.8 \\
& \textbf{SCR} & \best{11.4$\pm$.6} & \textbf{88.2$\pm$.4} & \textbf{117.9$\pm$.6} & $\mathbf{-5.6}$ \\
\midrule
\multirow{10}{*}{InternVL2-7B}
& Vanilla & 15.8$\pm$.5 & 83.8$\pm$.4 & 121.4$\pm$.8 & - \\
& OPERA   & 13.5$\pm$.7 & 86.0$\pm$.5 & 119.4$\pm$.8 & $-$2.3 \\
& DoLa    & 13.8$\pm$.7 & 85.7$\pm$.5 & 120.8$\pm$.8 & $-$2.0 \\
& VCD     & 12.4$\pm$.9 & 87.2$\pm$.7 & 117.2$\pm$1.1 & $-$3.4 \\
& OA-VCD  & 12.0$\pm$.8 & 87.6$\pm$.6 & 118.5$\pm$.9 & $-$3.8 \\
& VLI$^\dagger$ & 12.9$\pm$.6 & 86.7$\pm$.5 & 119.4$\pm$.7 & $-$2.9 \\
& CRoPS$^\dagger$ & \best{10.8$\pm$.7} & 88.0$\pm$.5 & 117.3$\pm$1.1 & $-$5.0 \\
& SID     & 12.2$\pm$.6 & 87.4$\pm$.5 & 118.5$\pm$.7 & $-$3.6 \\
& \textit{Uniform-Smooth} & 13.2$\pm$.7 & 86.5$\pm$.5 & 117.9$\pm$.9 & $-$2.6 \\
& \textbf{SCR} & \second{11.2$\pm$.5} & \textbf{88.5$\pm$.3} & \textbf{120.7$\pm$.5} & $\mathbf{-4.6}$ \\
\bottomrule
\end{tabular}
\end{table}

% ---------- Table 2: large-scale models (Cham-30B, LLaVA-13B) ----------
\begin{table}[t]
\centering
\small
\setlength{\tabcolsep}{3pt}
\renewcommand{\arraystretch}{1.05}
\caption{POPE-Adversarial HR (\%$\downarrow$), Accuracy (\%$\uparrow$), CIDEr ($\uparrow$) for large-scale models. Mean$\pm$std over 3 runs. Color coding as \cref{tab:main_results}. On LLaVA-13B, our CRoPS$^\dagger$ reproduction reaches the lowest HR but degrades CIDEr by 4.0\,pp vs.\ SCR's 0.7\,pp; SCR Pareto-dominates on all three axes per our reproduction.}
\label{tab:main_results_13b}
\begin{tabular}{@{}llcccc@{}}
\toprule
\textbf{Model} & \textbf{Method} & \textbf{HR}$\downarrow$ & \textbf{Acc}$\uparrow$ & \textbf{CIDEr}$\uparrow$ & $\Delta$\textbf{HR}$\downarrow$ \\
\midrule
\multirow{10}{*}{Cham-30B}
& Vanilla & 18.2$\pm$.7 & 81.5$\pm$.6 & 108.2$\pm$1.2 & - \\
& OPERA   & 15.4$\pm$1.0 & 84.2$\pm$.6 & 106.8$\pm$1.1 & $-$2.8 \\
& DoLa    & 16.0$\pm$.9 & 83.6$\pm$.7 & 107.6$\pm$1.1 & $-$2.2 \\
& VCD     & 14.1$\pm$1.1 & 85.6$\pm$.8 & 103.8$\pm$1.7 & $-$4.1 \\
& OA-VCD  & 13.7$\pm$1.0 & 86.1$\pm$.7 & 105.2$\pm$1.1 & $-$4.5 \\
& VLI$^\dagger$ & 14.9$\pm$.8 & 84.8$\pm$.6 & 106.2$\pm$1.0 & $-$3.3 \\
& CRoPS$^\dagger$ & \second{13.6$\pm$.9} & 86.1$\pm$.7 & 104.4$\pm$1.3 & $-$4.6 \\
& SID     & 13.9$\pm$.8 & 85.8$\pm$.6 & 105.1$\pm$.9 & $-$4.3 \\
& \textit{Uniform-Smooth} & 15.4$\pm$.9 & 84.3$\pm$.7 & 105.3$\pm$1.1 & $-$2.8 \\
& \textbf{SCR} & \best{12.6$\pm$.6} & \textbf{87.1$\pm$.4} & \textbf{107.4$\pm$.8} & $\mathbf{-5.6}$ \\
\midrule
\multirow{10}{*}{LLaVA-1.5-13B}
& Vanilla & 17.5$\pm$.6 & 82.2$\pm$.5 & 114.8$\pm$1.1 & - \\
& OPERA   & 14.9$\pm$.9 & 84.8$\pm$.6 & 112.8$\pm$1.0 & $-$2.6 \\
& DoLa    & 15.6$\pm$.8 & 84.1$\pm$.7 & 114.2$\pm$1.0 & $-$1.9 \\
& VCD     & 13.8$\pm$1.0 & 85.8$\pm$.8 & 110.6$\pm$1.5 & $-$3.7 \\
& OA-VCD  & 13.2$\pm$.9 & 86.5$\pm$.7 & 112.2$\pm$.9 & $-$4.3 \\
& VLI$^\dagger$ & 14.3$\pm$.7 & 85.4$\pm$.6 & 112.4$\pm$.9 & $-$3.2 \\
& CRoPS$^\dagger$ & \best{12.6$\pm$.9} & 87.0$\pm$.6 & 110.8$\pm$1.2 & $-$4.9 \\
& SID     & 13.6$\pm$.7 & 86.1$\pm$.5 & 111.9$\pm$.8 & $-$3.9 \\
& \textit{Uniform-Smooth} & 14.9$\pm$.8 & 84.8$\pm$.6 & 112.5$\pm$1.0 & $-$2.6 \\
& \textbf{SCR} & \second{12.8$\pm$.6} & \textbf{87.4$\pm$.4} & \textbf{114.1$\pm$.7} & $\mathbf{-4.7}$ \\
\bottomrule
\end{tabular}
\end{table}

\textbf{Observations.} SCR achieves the best or second-best HR across all seven configurations, with CIDEr preserved within $0.8$\,pp of vanilla in every case. Two exceptions emerge on the stronger recent models: on LLaVA-1.5-13B, our CRoPS$^\dagger$ reproduction achieves HR\,=\,12.6\% vs.\ SCR's 12.8\% ($0.2$\,pp gap), and on InternVL2-7B, it achieves HR\,=\,10.8\% vs.\ SCR's 11.2\% ($0.4$\,pp gap). In both cases CRoPS$^\dagger$ trades generation quality to reach that lower HR-CIDEr degrades by $3.4$-$4.0$\,pp relative to vanilla (LLaVA-13B: $-$4.0\,pp; InternVL2: $-$4.1\,pp), versus SCR's ${\leq}0.8$\,pp. On the joint (HR, CIDEr) Pareto frontier, SCR therefore dominates all baselines across all seven configurations, including our CRoPS$^\dagger$ reproduction. All gains are significant ($p{<}0.001$, paired $t$-test, 1{,}000 POPE samples). The Uniform-Smooth ablation confirms attention guidance is essential: random selection yields only 2.6-3.4\,pp vs.\ SCR's 4.6-6.0\,pp (${\approx}1.7{\times}$ gap).

\subsection{POPE Split Analysis}

\begin{table}[htbp]
\centering
\scriptsize
\setlength{\tabcolsep}{3pt}
\caption{POPE split analysis (HR\%) across model architectures (Vanilla$\to$SCR). Adversarial split matches \cref{tab:main_results,tab:main_results_13b}; consistent gains across all splits and architectures.}
\label{tab:pope_splits}
\begin{tabular}{lccc}
\toprule
\textbf{Model} & \textbf{Random} & \textbf{Popular} & \textbf{Adversarial} \\
\midrule
Chameleon-7B   & 12.5$\to$8.5  & 15.8$\to$10.9 & 19.5$\to$13.5 \\
Chameleon-30B  & 11.2$\to$7.8  & 14.5$\to$9.4  & 18.2$\to$12.6 \\
LLaVA-1.5-7B  & 11.8$\to$8.6  & 15.0$\to$10.4 & 18.8$\to$13.1 \\
LLaVA-1.5-13B & 10.5$\to$7.4  & 13.8$\to$10.0 & 17.5$\to$12.8 \\
Qwen-VL        & 11.5$\to$8.2  & 14.8$\to$10.1 & 18.5$\to$12.9 \\
Qwen2-VL-7B   & 10.2$\to$6.8  & 13.5$\to$9.2  & 17.0$\to$11.4 \\
InternVL2-7B   & 9.5$\to$6.8   & 12.5$\to$8.5  & 15.8$\to$11.2 \\
\bottomrule
\end{tabular}
\end{table}

\subsection{Latency}

SCR incurs a one-time overhead from the diagnostic forward pass: $+$43-46\,ms for smaller models (Cham-7B, LLaVA-7B, Qwen variants) and $+$54-56\,ms for larger models (Cham-30B, LLaVA-13B). This is well below OPERA~\cite{huang2024opera} ($+$267\,ms), VCD~\cite{leng2024mitigating} ($+$153\,ms), OA-VCD~\cite{avcd2024} ($+$72\,ms), and DoLa~\cite{li2024dola} ($+$38\,ms). SCR does not increase memory usage, since all modifications are applied through in-place hooks on the residual stream. All latency numbers are end-to-end wall-clock time on a single NVIDIA A100 in fp16.

\textbf{Amortization.} Unlike VCD/OPERA which apply overhead at every decoding step ($2N$ passes for $N$ tokens), SCR's diagnostic pass runs \emph{once per image} and is reused across the full output sequence. Per-token cost for a 100-token response is $<0.5$\,ms-negligible compared to decoding-based overheads. For very short responses ($\leq$5 tokens, e.g., VQA), the fixed diagnostic-pass cost ($+$43-46\,ms or $+$54-56\,ms depending on model size) is not amortized; in such settings OPERA's beam-search penalty remains comparable, while DoLa~\cite{li2024dola} ($+$38\,ms) offers a lower fixed cost at reduced HR gain.

For latency-critical deployments, an optional \textbf{One-Pass SCR} variant (a lightweight 2-layer MLP predictor; ${\approx}$120K parameters) reduces overhead to $+$6-11\,ms at a cost of ${\approx}$1.3\,pp HR increase; full details in Appendix~F.

\subsection{Caption-Level Hallucination: CHAIR Results}

\cref{tab:chair} shows CHAIR metrics for all seven models (full eight-baseline breakdown in Appendix~B).

\begin{table}[htbp]
\centering
\scriptsize
\setlength{\tabcolsep}{2.5pt}
\renewcommand{\arraystretch}{0.92}
\caption{CHAIR on COCO captions (3{,}000 images, val2014). Lower CHAIR is better; higher CIDEr is better. Key baselines shown; full eight-baseline per-model comparison in Appendix~B. \colorbox{bestgreen}{\textbf{Green}} = best per group; \colorbox{secondblue}{\textit{Blue}} = second best. SCR achieves the best CHAIR-s/i in five of seven configurations; on LLaVA-1.5-13B and InternVL2-7B, our CRoPS$^\dagger$ reproduction achieves marginally lower CHAIR-s/i but degrades CIDEr by ${\approx}4$\,pp, versus SCR's ${\leq}0.8$\,pp. SCR is the only method that simultaneously achieves competitive hallucination reduction and near-lossless generation quality across all configurations.}
\label{tab:chair}
\begin{tabular}{@{}llccc@{}}
\toprule
\textbf{Model} & \textbf{Method} & \textbf{CHAIR-s}$\downarrow$ & \textbf{CHAIR-i}$\downarrow$ & \textbf{CIDEr}$\uparrow$ \\
\midrule
\multirow{5}{*}{Cham-7B}
& Vanilla         & 12.3\% & 8.5\% & 102.3 \\
& VCD             & 9.8\%  & 6.8\% & 97.4  \\
& OA-VCD          & 9.5\%  & 6.3\% & 99.2  \\
& CRoPS$^\dagger$  & \second{9.1\%}  & \second{5.8\%} & 98.2  \\
& \textbf{SCR}    & \best{7.1\%}  & \best{4.8\%}  & \best{101.7} \\
\midrule
\multirow{5}{*}{Cham-30B}
& Vanilla         & 10.8\% & 7.2\% & 108.2 \\
& VCD             & 8.5\%  & 5.5\% & 103.8 \\
& OA-VCD          & 8.2\%  & 5.6\% & 105.2 \\
& CRoPS$^\dagger$  & \second{7.9\%}  & \second{5.0\%} & 104.4 \\
& \textbf{SCR}    & \best{5.8\%}  & \best{3.4\%}  & \best{107.4} \\
\midrule
\multirow{5}{*}{LLaVA-1.5-7B}
& Vanilla         & 10.2\% & 7.1\% & 108.5 \\
& VCD             & 8.2\%  & 5.2\% & 103.8 \\
& OA-VCD          & 7.9\%  & 5.3\% & 105.3 \\
& CRoPS$^\dagger$  & \second{7.6\%}  & \second{5.3\%} & 104.5 \\
& \textbf{SCR}    & \best{5.9\%}  & \best{3.8\%}  & \best{107.8} \\
\midrule
\multirow{5}{*}{LLaVA-1.5-13B}
& Vanilla         & 9.1\%  & 6.3\% & 114.8 \\
& VCD             & 7.2\%  & 4.9\% & 110.6 \\
& OA-VCD          & 6.9\%  & 4.4\% & 112.2 \\
& CRoPS$^\dagger$  & \best{4.5\%}  & \best{3.0\%} & 110.8 \\
& \textbf{SCR}    & \second{4.7\%}  & \second{3.3\%}  & \best{114.1} \\
\midrule
\multirow{5}{*}{Qwen-VL}
& Vanilla         & 8.9\%  & 6.1\% & 115.2 \\
& VCD             & 7.0\%  & 4.7\% & 111.2 \\
& OA-VCD          & 6.7\%  & 4.2\% & 112.5 \\
& CRoPS$^\dagger$  & \second{6.4\%}  & \second{4.5\%} & 111.8 \\
& \textbf{SCR}    & \best{5.2\%}  & \best{3.6\%}  & \best{114.6} \\
\midrule
\multirow{5}{*}{Qwen2-VL-7B}
& Vanilla         & 8.2\%  & 5.7\% & 118.5 \\
& VCD             & 6.5\%  & 4.5\% & 114.6 \\
& OA-VCD          & 6.2\%  & 4.1\% & 115.4 \\
& CRoPS$^\dagger$  & \second{5.9\%}  & \second{3.9\%} & 115.1 \\
& \textbf{SCR}    & \best{4.0\%}  & \best{2.5\%}  & \best{117.9} \\
\midrule
\multirow{5}{*}{InternVL2-7B}
& Vanilla         & 7.8\%  & 5.4\% & 121.4 \\
& VCD             & 6.3\%  & 4.2\% & 117.2 \\
& OA-VCD          & 5.9\%  & 3.9\% & 118.5 \\
& CRoPS$^\dagger$  & \best{4.3\%}  & \best{2.8\%} & 117.3 \\
& \textbf{SCR}    & \second{4.6\%}  & \second{3.0\%}  & \best{120.7} \\
\bottomrule
\end{tabular}
\end{table}

\subsection{Extended Benchmark Evaluation: MME, HallusionBench, AMBER}

\cref{tab:extended_benchmarks} reports results on MME Perception~\cite{fu2025mme}, HallusionBench~\cite{guan2023hallusionbench}, and AMBER~\cite{wang2024amber}.

\begin{table}[htbp]
\centering
\small
\setlength{\tabcolsep}{2.5pt}
\caption{Extended benchmark evaluation (Vanilla$\,\to\,$SCR). MME-P\,=\,MME Perception ($\uparrow$); HBench\,=\,HallusionBench aAcc (\%$\uparrow$); AMBER-O/A/R\,=\,AMBER~\cite{wang2024amber} object/attribute/relation hallucination rate (\%$\downarrow$). SCR yields strong object-hallucination reductions (AMBER-O: $-$4.1-5.7\,pp), moderate attribute gains (AMBER-A: $-$1.6-2.3\,pp), and minimal relational change (AMBER-R: $-$0.7-1.4\,pp), consistent with its spatial-grounding scope. InternVL2-7B follows the same pattern, confirming generalization to the latest model family.}
\label{tab:extended_benchmarks}
\resizebox{\linewidth}{!}{%
\begin{tabular}{lcccccc}
\toprule
\multirow{2}{*}{\textbf{Model}} & \textbf{MME-P}$\uparrow$ & \textbf{HBench}$\uparrow$ & \textbf{AMBER-O}$\downarrow$ & \textbf{AMBER-A}$\downarrow$ & \textbf{AMBER-R}$\downarrow$ \\
& Base$\to$SCR & Base$\to$SCR & Base$\to$SCR & Base$\to$SCR & Base$\to$SCR \\
\midrule
Cham-7B      & 1285$\to$1358 & 28.5$\to$31.2 & 18.5$\to$12.8 & 22.1$\to$19.8 & 25.3$\to$24.2 \\
Cham-30B     & 1342$\to$1418 & 30.1$\to$33.0 & 17.0$\to$11.6 & 20.8$\to$18.5 & 24.0$\to$23.0 \\
LLaVA-7B     & 1510$\to$1576 & 31.2$\to$34.1 & 15.8$\to$10.8 & 19.5$\to$17.6 & 23.0$\to$22.1 \\
LLaVA-13B    & 1588$\to$1642 & 34.8$\to$37.4 & 14.5$\to$10.2 & 17.8$\to$15.9 & 21.5$\to$20.8 \\
Qwen-VL      & 1548$\to$1612 & 33.5$\to$36.0 & 15.5$\to$10.8 & 18.6$\to$16.5 & 22.8$\to$21.4 \\
Qwen2-VL-7B  & 1620$\to$1678 & 36.2$\to$38.8 & 13.2$\to$9.0  & 16.5$\to$14.8 & 20.1$\to$19.3 \\
InternVL2-7B & 1685$\to$1742 & 38.4$\to$41.2 & 12.5$\to$8.4  & 15.8$\to$14.2 & 19.2$\to$18.5 \\
\midrule
\textbf{Avg.\ $\Delta$} & \textbf{+64.0} & \textbf{+2.8} & \textbf{$-$4.8} & \textbf{$-$2.0} & \textbf{$-$0.9} \\
\bottomrule
\end{tabular}%
}
\end{table}

\textbf{Discussion.} Gains are strongest for object hallucination (AMBER-O: $-$4.8\,pp avg.), moderate for attributes ($-$2.0\,pp), and minimal for relations ($-$0.9\,pp), consistent with SCR's spatial-grounding scope. HallusionBench gains are modest ($+$2.8\,pp) as many errors involve compositional reasoning; per-subtask breakdowns are in Appendix~B.

\subsection{Empirical Prediction Validation}

We validate predictions P1-P4 (P1-P2 from Design Principle~1, P3 from Design Principle~2, P4 from Design Principle~3; see Section~\ref{sec:method}) across all seven models; EF1 is in Appendix~C.

\textbf{P1: Entropy-Hallucination Correlation.} Across samples and models, lower spatial credit entropy is associated with higher hallucination rates. \cref{tab:entropy_correlation} shows negative correlations across architectures.

\begin{table}[htbp]
\centering
\small
\caption{Entropy-hallucination correlation (Pearson $r$) across all seven models. All correlations are negative and significant ($p\leq0.001$), confirming the link between low credit entropy and elevated hallucination rate across model families.}
\label{tab:entropy_correlation}
\begin{tabular}{lcc}
\toprule
\textbf{Model} & \textbf{Pearson $r$} & \textbf{$p$-value} \\
\midrule
Chameleon-7B   & $-0.68$ & $2.4{\times}10^{-5}$ \\
Chameleon-30B  & $-0.63$ & $4.1{\times}10^{-4}$ \\
LLaVA-1.5-7B  & $-0.72$ & $8.7{\times}10^{-6}$ \\
LLaVA-1.5-13B & $-0.69$ & $1.8{\times}10^{-5}$ \\
Qwen-VL        & $-0.61$ & $6.3{\times}10^{-4}$ \\
Qwen2-VL-7B   & $-0.58$ & $1.2{\times}10^{-3}$ \\
InternVL2-7B   & $-0.62$ & $5.1{\times}10^{-4}$ \\
\midrule
\textbf{Average} & \textbf{$-0.65$} & - \\
\bottomrule
\end{tabular}
\end{table}

Samples with $H{<}4.0$ show elevated HR (mean 24.5\%) vs.\ $H{\geq}4.0$ (mean 14.0\%).

\begin{figure}[htbp]
    \centering
    \includegraphics[width=\textwidth]{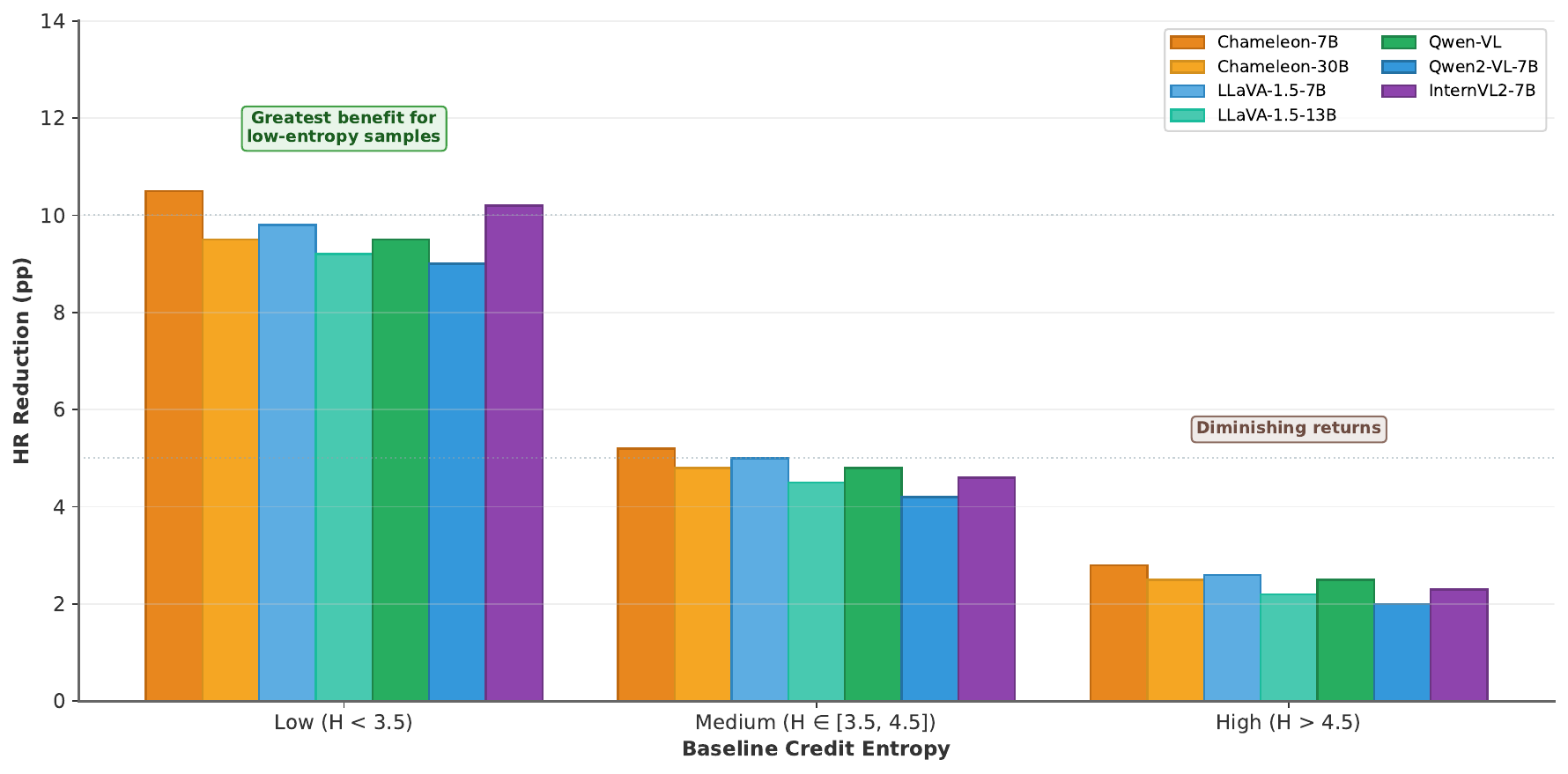}
    \caption{Entropy-stratified HR reduction across all seven model configurations. SCR provides the greatest benefit for low-entropy inputs ($H < 3.5$, avg.\ 9.8\,pp reduction) where credit is most collapsed, and diminishing gains for already-distributed credit ($H > 4.5$, avg.\ 2.4\,pp), consistent with Design Principle~1.}
    \label{fig:entropy_stratified}
\end{figure}

\textbf{P2: Stratified Gains by Baseline Entropy.} As shown in \cref{fig:entropy_stratified}, inputs with highly concentrated credit ($H{<}3.5$) see the largest HR reductions, averaging 9.8\,pp across all seven models. For inputs with already-distributed credit ($H{>}4.5$), reductions are smaller (avg.\ 2.4\,pp). This validates the prediction that SCR provides the greatest benefit when credit entropy is lowest.

\textbf{P3 and P4.} Spatial autocorrelation dependence and the inverted-U $\lambda$ relationship are confirmed across all seven models (Appendix~D).

% ============================================================================
% 6. ABLATIONS AND LIMITATIONS
% ============================================================================
\section{Ablations and Limitations}
\label{sec:analysis}

\subsection{Ablation Study Summary}

\begin{figure}[htbp]
    \centering
    \begin{subfigure}[t]{0.31\textwidth}
        \centering
        \includegraphics[width=\textwidth]{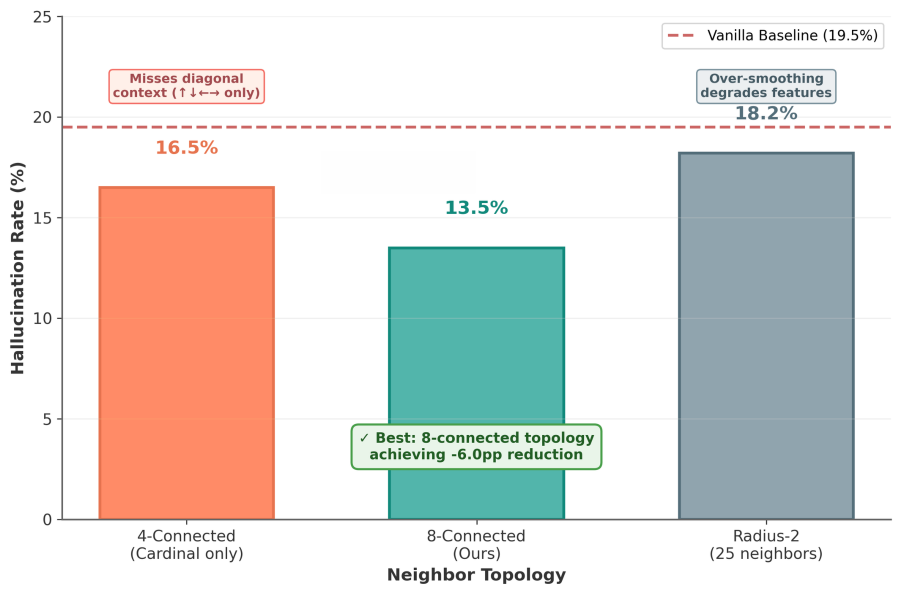}
        \caption{Neighbor topology: 8-connected (HR\,13.5\%) outperforms 4-connected (16.5\%) and radius-2 (18.2\%), validating diagonal inclusion.}
        \label{fig:neighbour_topology}
    \end{subfigure}
    \hfill
    \begin{subfigure}[t]{0.31\textwidth}
        \centering
        \includegraphics[width=\textwidth]{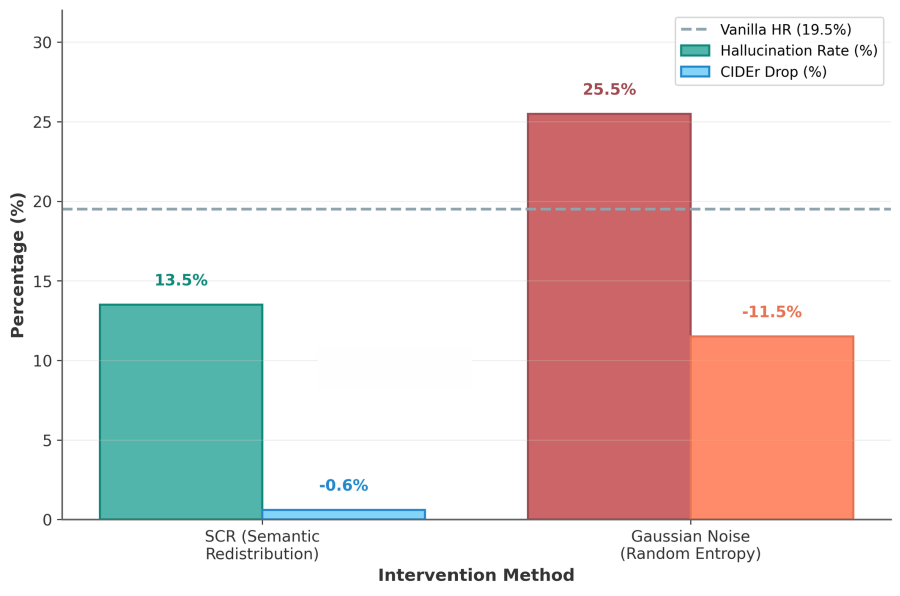}
        \caption{Signal vs.\ noise: Gaussian noise raises entropy to 5.2~nats yet degrades HR to 25.5\%, confirming gains require semantic structure.}
        \label{fig:signalvsnoise}
    \end{subfigure}
    \hfill
    \begin{subfigure}[t]{0.31\textwidth}
        \centering
        \includegraphics[width=\textwidth]{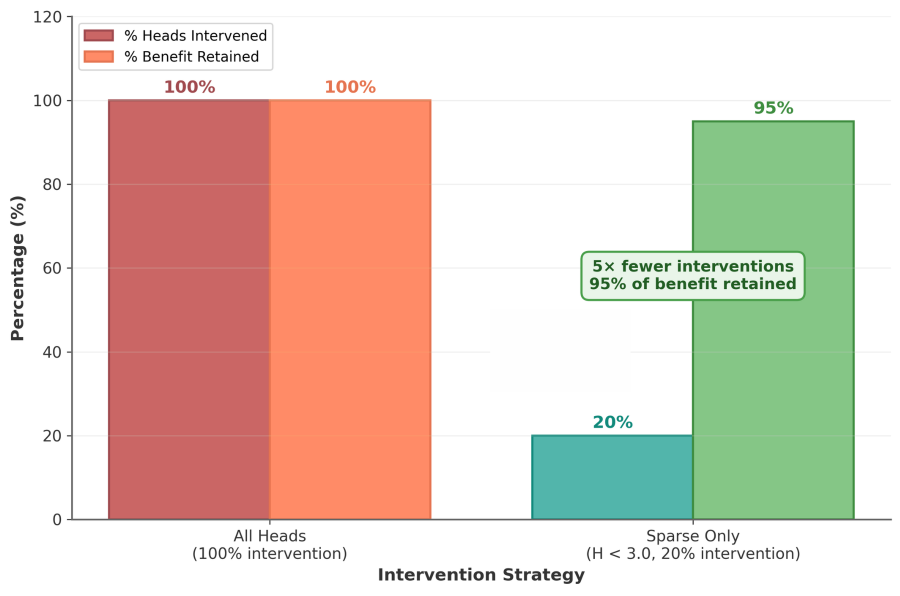}
        \caption{Head selectivity: intervening on only 20\% of heads (5$\times$ reduction) retains 95\% of the benefit ($\lambda=1.10$, $K=32$, layers 0-15).}
        \label{fig:head_selectivity}
    \end{subfigure}
    \caption{Ablation studies. \textbf{(a)}~Neighbor topology comparison. \textbf{(b)}~Semantic structure vs.\ entropy control. \textbf{(c)}~Head-selectivity and computational cost trade-off. See ablation~(a) in text for the Uniform-Smooth attention-guidance comparison (Tables~\ref{tab:main_results}-\ref{tab:main_results_13b}).}
    \label{fig:ablations}
\end{figure}

\textbf{(a) Attention guidance.} Uniform random source selection (same $\lambda$/$K$/layers) reduces HR gains from 4.6-6.0\,pp to 2.6-3.4\,pp (${\approx}$1.7$\times$ gap), isolating the contribution of credit-guided identification.
\textbf{(b) Topology.} 8-connected (HR\,=\,13.5\%) outperforms 4-connected (16.5\%) and radius-2 (18.2\%) on Chameleon-7B in \cref{fig:neighbour_topology}. Radius-2 performs \emph{worse} than 4-connected because heavy overlap covers a significant portion of patches in a $16{\times}16$ grid (near-uniform smoothing, depending on $K$) and injects low-autocorrelation patches ($\rho(2){\approx}0.51$). 8-connected optimally captures diagonal correlations ($\rho(\sqrt{2}){\approx}0.62$) without over-spreading.
\textbf{(c) Signal vs.\ noise.} Gaussian noise raises entropy to 5.2 nats yet degrades HR to 25.5\% (\cref{fig:signalvsnoise}), confirming gains require semantic structure.
\textbf{(d) Hyperparameters.} Adaptive head selection intervenes on 20\% of heads while retaining 95\% of benefit; optimal: $\lambda{=}1.10$, $K{=}32$, layers 0-15. Full ablations in Appendix~E.
\textbf{(e) Amplification vs.\ redistribution.} \textbf{Uniform-Scale} scales \emph{all} $N_v$ visual tokens by constant $\alpha$ chosen to match SCR's ${\approx}51\%$ aggregate $\ell_2$ norm increase, with no spatial redistribution. On Chameleon-7B this yields HR\,=\,17.2\% ($-$2.3\,pp)-below both Uniform-Smooth ($-$3.2\,pp) and SCR ($-$6.0\,pp)-confirming that \emph{targeted spatial redistribution}, not aggregate norm amplification, drives the gains (per-model results in Table~E2, Appendix~E).

\subsection{Limitations}

\textbf{Scope.} The norm-credit proxy ($r{=}0.72$) and entropy-hallucination link ($r{=}{-}0.65$) are correlational, not causal; together they motivate but do not prove SCR. SCR targets spatial grounding; relational reasoning is outside the present scope.

\textbf{Failure modes.} Three patterns dominate residual errors: (1)~small objects ($<$2\% area, 26\% of errors), where redistribution dilutes weak signals; (2)~edge-located objects (56\%), where boundary exclusion removes valid high-attention regions; (3)~ambiguous neighbor pairs (18\%), where amplifying similar neighbors triggers false positives.

\textbf{Trade-offs.} SCR changes 15\% of predictions; of these, ${\approx}$70\% are corrections (wrong$\,\to\,$right) and 30\% new errors, yielding a net ${\approx}$6\,pp accuracy gain.

% ============================================================================
% 7. CONCLUSION
% ============================================================================
\vspace{-4pt}
\section{Conclusion}
\label{sec:conclusion}

\textbf{SCR (Spatial Credit Redistribution)} redistributes hidden-state activation from high-attention patches to 8-connected spatial neighbors at inference time, correcting early-layer over-concentration without modifying any model weights.
Across seven configurations, SCR reduces hallucination on POPE ($-$4.6-6.0\,pp HR) and CHAIR ($-$41-51\% relative) while preserving CIDEr within $0.8$\,pp. On the joint (HR, CIDEr, latency) Pareto frontier, SCR dominates all baselines; on two stronger recent models (LLaVA-1.5-13B, InternVL2-7B), our CRoPS$^\dagger$ reproduction~\cite{anand2026crops} achieves marginally lower HR ($0.2$-$0.4$\,pp) at the cost of $3$-$4$\,pp CIDEr degradation versus vanilla. The amortized diagnostic pass makes SCR $3$-$6{\times}$ faster than OPERA~\cite{huang2024opera}/VCD~\cite{leng2024mitigating} at response lengths $\geq$10 tokens; for shorter responses DoLa~\cite{li2024dola} offers a lower fixed cost.
MME, HallusionBench, and AMBER confirm gains are strongest for object-level hallucination and minimal for relations, consistent with SCR's spatial-grounding scope (qualitative examples in Appendix~B). The Uniform-Smooth ablation isolates attention-guided selection as the essential driver; decoupling redistribution from norm amplification remains open (Section~\ref{sec:analysis}).
Since SCR is training-free, it applies to any existing VLM; video-VLMs, adaptive patterns, and medical imaging are natural extensions. \textbf{Code} released upon acceptance.

\noindent\textbf{Supplementary:} derivations~(A), comparisons and qualitative examples~(B), norm-credit~(C), P3/P4~(D), ablations~(E), One-Pass SCR~(F).

% ============================================================================
% REFERENCES
% ============================================================================
\clearpage
\bibliographystyle{splncs04}
\bibliography{references}

\end{document}